\documentclass[10pt,journal,compsoc]{IEEEtran}
\ifCLASSOPTIONcompsoc
  \usepackage[nocompress]{cite}
\else
  \usepackage{cite}
\fi
\ifCLASSINFOpdf
\else
\fi

\usepackage{color}
\usepackage{caption}
\usepackage[hyphens]{url}  
\usepackage{graphicx} 
\usepackage{amsmath,amssymb}
\usepackage{adjustbox}

\begin{document}
%
\title{One-Shot Adaptation of GAN in Just One CLIP}
%
%
%
%

\author{Gihyun~Kwon,
        and~Jong~Chul~Ye,~\IEEEmembership{Fellow,~IEEE}
\IEEEcompsocitemizethanks{\IEEEcompsocthanksitem Gihyun Kwon is with the Department of Bio and Brain Engineering, Korea Advanced Institute of Science and Technology (KAIST), Daejeon 34141, South Korea.\protect\\
E-mail: cyclomon@kaist.ac.kr
\IEEEcompsocthanksitem Jong Chul Ye is with the Graduate School of AI, Korea Advanced Institute of Science and Technology (KAIST), Daejeon 34141, South Korea.

E-mail: jong.ye@kaist.ac.kr
}
\thanks{Manuscript received December 5, 2022;}}

%
%

\markboth{Journal of \LaTeX\ Class Files,~Vol.~14, No.~8, August~2015}%
{Shell \MakeLowercase{\textit{et al.}}: Bare Demo of IEEEtran.cls for Computer Society Journals}
%



\IEEEtitleabstractindextext{
\begin{abstract}
There are many recent research efforts to fine-tune a pre-trained generator with a few target images to generate images of a novel domain. 
Unfortunately, these methods often suffer from overfitting or under-fitting when fine-tuned with a single target image.
To address this, here we present  a novel single-shot GAN adaptation method  through  unified CLIP space manipulations.
Specifically, our model employs a two-step training strategy: reference image search in the source generator using a CLIP-guided latent optimization, followed by generator fine-tuning with a novel loss function that imposes CLIP space  consistency between the source and adapted generators. To further improve the adapted model to produce spatially consistent samples with respect to the source generator, we also propose contrastive regularization for patchwise relationships in the  CLIP space.
Experimental results show that our model generates diverse outputs with the target texture and  outperforms the baseline models both qualitatively and quantitatively.
Furthermore, we show that our CLIP space manipulation strategy allows more effective attribute editing. Our Github source is as follows: \url{https://github.com/cyclomon/OneshotCLIP}
\end{abstract}

}

\maketitle

\IEEEdisplaynontitleabstractindextext

%
\IEEEpeerreviewmaketitle

\IEEEraisesectionheading{\section{Introduction}\label{sec:introduction}}

%
%
%
%
\IEEEPARstart{R}{ecently}, several studies have tried to fine-tune a pre-trained generator model with limited number of target images so that the fine-tuned model can generate images of a novel domain. 
Early methods \cite{TGAN,freezeD,BSA} showed the results using fine-tuned the models with about 100 training images. Subsequent studies \cite{Fewshot,EWC} have shown that GAN domain adaptation is possible even in extreme situations using fewer than 10 training images. A recent method \cite{mind} attempted to transform a pre-trained source generator to generate novel domain images by fine-tuning with only one target image. 

Although above methods show good performance in limited-shot situations, when there are smaller number of available training images, the generation performance drops significantly due to the severe overfitting. The problem is more pronounced when only one target image is available, which is the situation we are trying to solve.
Although a recently proposed model  \cite{mind} has mitigated the overfitting problem, the model often suffers from underfitting so that the model cannot fully reflect the domain of the target image.

\begin{figure}[t!]
\centering
\includegraphics[width=\linewidth]{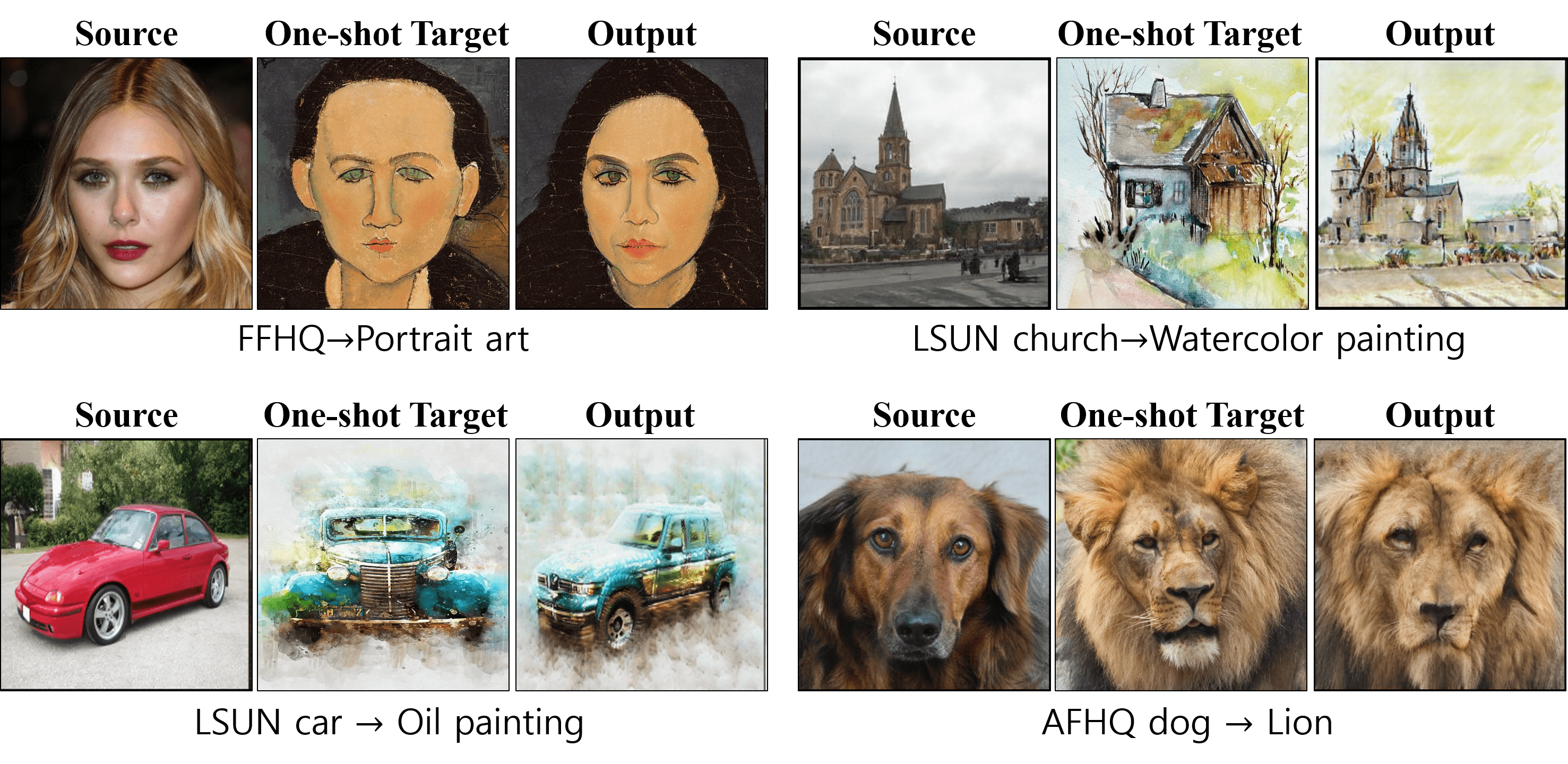}
\caption{Various domain adaptation results from our model. Our model successfully fine-tuned the models pre-trained on large data into a target domain with only a single shot target image. The adapted model can generate various images corresponding to the target domain while preserving the content attribute of the source domain. More generated samples can be found in the experiment section. }
\label{fig:first}
\end{figure}

To address this, here we present a novel single-shot fine-tuning approach of the pre-trained generator  by using a unified CLIP space manipulation, 
which shows better perceptual quality compared to other state-of-the-art few-shot adaptation models. 
For example, Fig. \ref{fig:first} shows that
our model successfully generates new images with only a single target by fine-tuning the source model trained on large datasets (e.g. FFHQ, LSUN church, LSUN cars, and AFHQ dog). One of the most important contributions of this work is the discovery of the importance of the unified CLIP space manipulation.


Specifically, our method consists of two steps: finding the reference image and fine-tuning the model, all in the CLIP space.
More specifically, inspired by the idea of the recent single-shot adaptation method called {Mind the GAP} \cite{mind}, the first step is to find the image from the pre-trained source generator that most closely resembles a given target image $I_{trg}$ and uses it as a reference image $I_{ref}$.
In particular,    unlike existing GAN inversion methods and Mind the GAP \cite{mind} which uses pixel- wise similarity, we propose a new strategy that precisely aligns the semantic attributes between $I_{ref}$ and  $I_{trg}$ 
   in the CLIP space.  
   In the second step, we fine-tune the model using the patch-wise and global discriminators. In this stage,
$I_{ref}$ obtained in the previous step plays an important role as a reference point that guides the adapted generator to maintain the principal attribute information (e.g. content shape) of the source generator. 
To further match the source-target domain consistency, we propose a new consistency loss that maintains the CLIP space relation between the generated samples. Also, in order to maintain detailed spatial information between source and target models, we propose to maintain patch-wise semantic relations with contrastive learning in the CLIP space.  To further verify the performance of our model, 
we show that our CLIP space manipulation strategy allows flexible and effective attribute editing. 

Our contribution can be summarized as follows:
\begin{itemize}
\item We proposed to use CLIP in the latent search stage to find a reference that is more suitable for model adaptation, which have not been tried in the previous work.
\item We proposed patch- and sample-wise consistency regularization in the CLIP embedding space, which is also the very \textbf{first} trial in GAN adaptation task.
\item Our method outperforms other limited-shot GAN adaptation methods in both of qualitative and quantitative results. 
\item Our method has flexible framework so that it can be easily applied to other tasks, such as attribute editing, text-guided adaptation, etc. 
\end{itemize}

\section{Related Work}
\subsection{Generative Adversarial Networks}
Recent advances in generative adversarial networks (GAN) has shown impressive performance  in generating realistic images. Starting from the seminal work of GAN \cite{gan}, Progressive GAN \cite{progan} and BigGAN \cite{biggan} models showed significantly improved generation performance for human face dataset and natural images. Furthermore, StyleGAN \cite{stylegan} showed generation quality that is almost indistinguishable from the real data. 

One of the great advantages of StyleGAN is the model flexibility, which has inspired numerous subsequent models through modifications of StyleGAN,
such as StyleGAN with a modified structure for light weight model~\cite{mobilestylegan}, a disentangled model with additional latent space by modifying the architecture of StyleGAN~\cite{sni,diagonal}, a model with additional encoder network~\cite{stylemap,alae}, etc.
Recently proposed StyleGAN2 \cite{stylegan2} showed further improved generation performance through more efficient model structure and training strategies.
StyleGAN-ADA \cite{ADA}, also another version of StyleGAN2, showed better generational performance through a novel training strategy using adaptive augmentation in discriminator training. Besides, various methods have been proposed to perform diverse tasks by leveraging the characteristics of StyleGAN,
such as semantic segmentation \cite{datasetgan,labels4free} for dataset generation, 3D image rendering \cite{stylerig,3dgen},  inverse problems \cite{pulse,ganprior}, etc.

\subsection{Image manipulation} 
Manipulating specific attribute or domain of a  given image is an important topics in  computer vision.  
CycleGAN \cite{cycleGAN} successfully converted the image to other domains with novel cycle consistency,
which was followed by several improved image translation methods \cite{gcgan,distanceGAN}. The recent contrastive unpaired translation model (CUT) \cite{cut} approach achieved state-of-the-art performance by preserving the spatial attributes between the input and the generated output by applying contrastive learning to the patchwise feature embedding space.

In addition, multi-domain models that enable arbitrary image transformation between various domains have also been developed. StarGAN~\cite{stargan} is the first work
for  multi-domain translation through a single common generator, and StarGANv2~\cite{starganv2} extended the multi-domain single-output translation capability
of StarGAN into multi-domain diverse image translation.

In addition, various attribute manipulation methods leveraging the disentangled characteristic of StyleGAN have been proposed lately. These include a method to find specific direction for independent attribute by separating the principal component of the latent space~\cite{ganspace}, or to find a latent space that can edit a meaningful area of generated images by additionally using label information~\cite{interface} or {semantic map~\cite{stylefusion}}. 
Since these StyleGAN-based processing methods inevitably require a mapping of a given image to StyleGAN's latent space, various StyleGAN inversion methods have also been proposed. These include methods which train additional encoder model suitable for StyleGAN \cite{e4e,psp}, 
or latent optimization methods  for a given image \cite{i2s,ii2s}, etc.

Recently, with the introduction of the CLIP model \cite{clip} by OpenAI, image manipulation methods through text conditions,
such as text-to-image generation~\cite{fusedream} and text-guided style transfer~\cite{clipstyler},
 have received a lot of attention.  In addition, by exploiting the disentangled characteristic of StyleGAN, StyleCLIP \cite{styleclip} proposed a method for editing the latent attribute of StyleGAN through text, and StyleGAN-NADA \cite{nada} proposed a method of adapting the model to a novel domain through texts.
 {
 Especially, StyleGAN-NADA further showed the versatility of model adaptation for a given target image.}

\subsection{Few-shot domain adaptation}
The purpose of adapting a generative model using few-shot images is to guide the model to a target domain while inheriting the diversity of source model trained on large-scale data. One of the first works of GAN adaptation is TransferGAN~\cite{TGAN}, which provided a novel insight of few-shot generation as a kind of transfer learning. Various subsequent few-shot GAN adaptation models focused on preventing the generator from overfitting by introducing regularization components during the fine-tuning process. FreezeD~\cite{freezeD} performed GAN adaptation by freezing some layers of the discriminator in order to maintain the prior features of generator model. Others proposed to mix the weights  of fine-tuned model and source model \cite{mixing}, or maintain singular vectors of the pre-trained weights~\cite{singular}. In addition, EWC~\cite{EWC} used Fisher information to identify important layers of generator features and freeze them to prevent overfitting.

Recently, Ojha et al \cite{Fewshot} proposed an improved GAN adaptation method even in few-shot setting of less than 10 training samples by exploiting the cross-domain consistency and patch-based adversarial training. In the next version, Xiao et al \cite{rssa} tried to solve the overfitting problem with spatial structural alignment framework. 
Furthermore, Mind the GAP~\cite{mind} attempted to fine-tune the generator model using only one target image. 
Specifically, by 
aligning the vector direction within CLIP-space, stable domain adaptation was possible \cite{mind}. 

Although recently proposed models addressed the overfitting problems of GAN adaptation, most of previous methods still suffer from overfitting when the available samples are extremely limited. 
Mind the GAP could solve the overfitting with single-shot domain adaptation, but we empirically discovered that this model suffer from underfitting problems when the target image textures are complex or the appearance is far from source domain. 

\begin{figure}[!hbt]
\centering
\includegraphics[width=\linewidth]{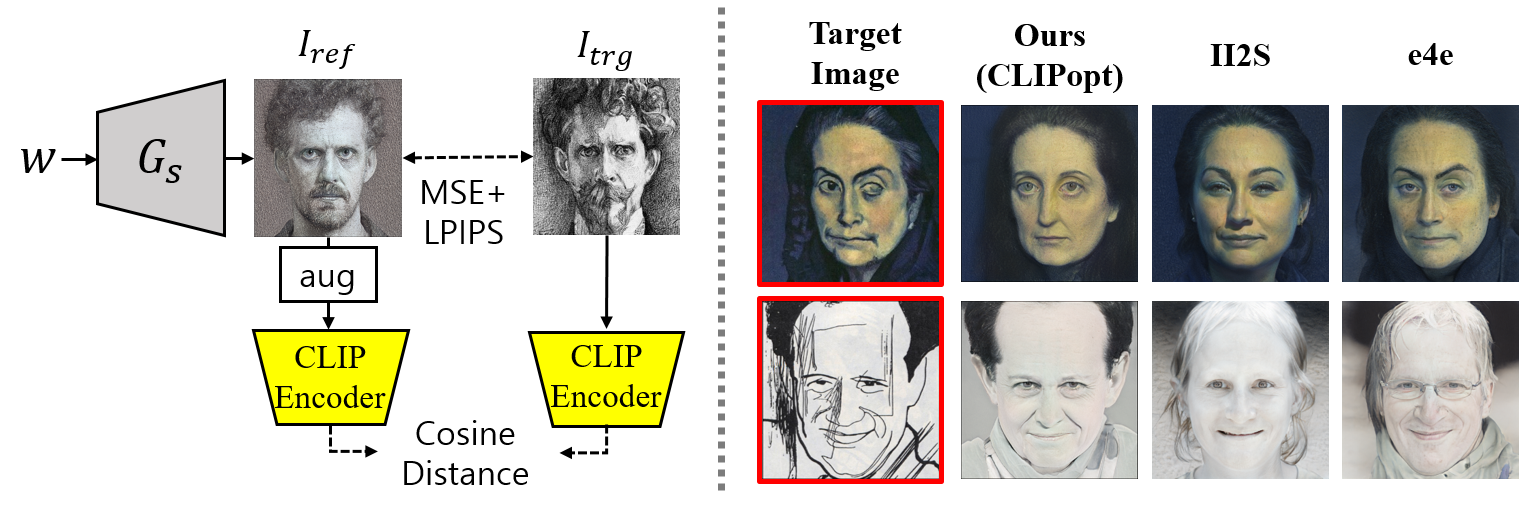}
\caption{(Left) Overview of our proposed CLIP-guided latent optimization to find the reference image $I_{ref}$. (Right) Comparison  between various baselines. Our results contain the desired attribute of the target image and outperforms the baseline results.}
\label{fig:opt}
\end{figure}
  \begin{figure}[!hbt]
\centering
\includegraphics[width=\linewidth]{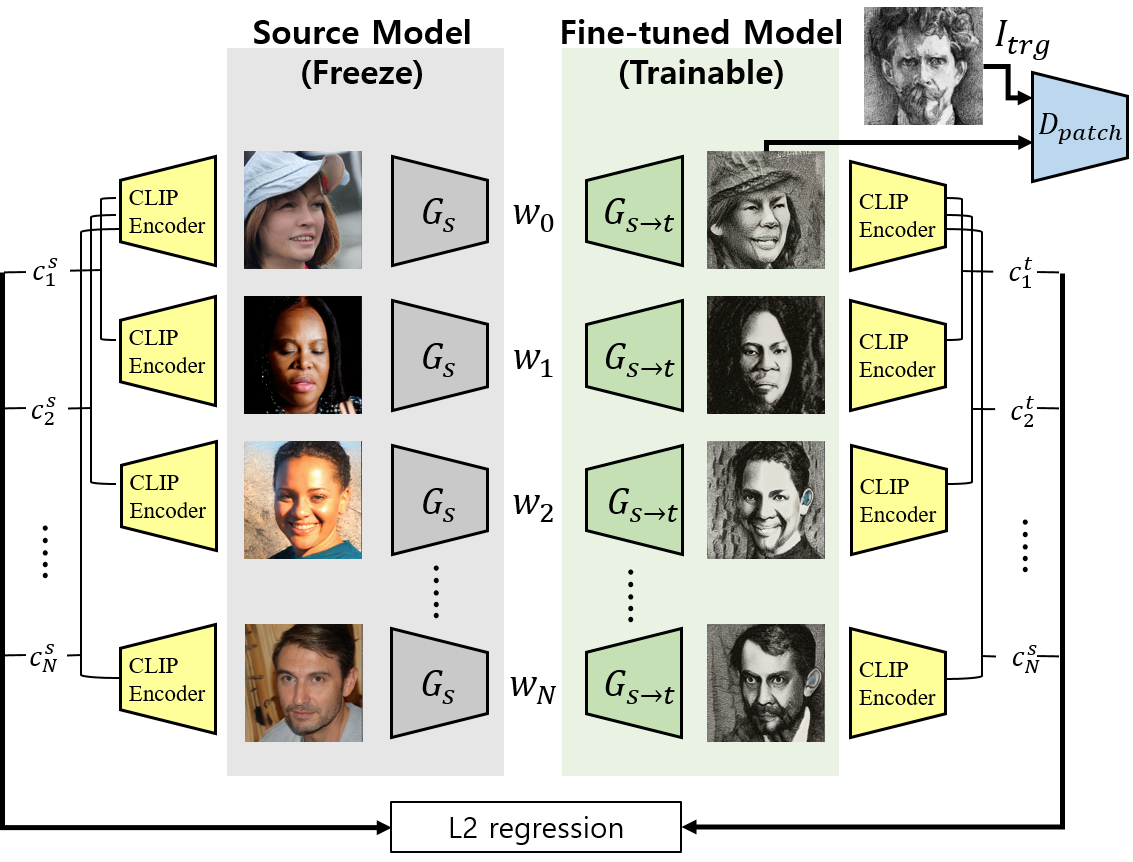}
\caption{Overview of our cross-domain semantic consistency in the CLIP space. We generate image from the source model $G_s$ and the target model $G_t$ with the same latents $w$. Then we obtain the embedded feature vector using a pre-trained CLIP model, 
and the cosine similarity scores between the embedded vectors in the CLIP space are calculated. The similarity scores between both domains ($c^t, c^s$) are aggregated through the $l_2$ regression. To guide the texture of the $G_t$ to match the target $I_{trg}$, we additionally use the patch discriminator.  }
\label{fig:method}
\end{figure}

\section{Methods}


By extending the existing approaches, our goal is to find 
  improved regularization and training strategies,
  which can adapt the model robustly without over- and under- fitting  problems regardless of the target image types or pre-trained source domains.
One of the most important contributions of this work is the discovery  of the importance of the unified CLIP space manipulation.

Specifically, our method consists of two steps. First, similar to the previous work \cite{mind}, we search a latent code $w_{ref}$ in the latent space of the source domain generator $G_{s}$ so that it can generate reference image  $I_{ref}=G(w_{ref})$ that is most similar to a given single-shot target image $I_{trg}$. 
However, in contrast to Mind the GAP \cite{mind} which uses pixel-wise similarity, we found that the CLIP space similarity significantly improves the search.
The 
resulting image $I_{ref}$ is then used as a reference point to be aligned with $I_{trg}$ in the next step of model fine-tuning. 
In the second step, we fine-tune the pre-trained generator to create a target generator $G_{t}$ which follow the domain information of $I_{trg}$. At this time, our goal is to guide $G_t$ to maintain the diverse content attribute of $G_s$ by aligning the target image $I_{trg}$ and the reference image $I_{ref}$ obtained in the previous step. Again we reinforce this by using regularizations in the CLIP space  to enforce  semantic consistency between source and target generators.
More detailed descriptions are as follows.

\subsection{Step 1: Clip-guided Latent Search}

In order to find the image most consistent with the target image, the previous adaptation model Mind the Gap~\cite{mind} leveraged the existing StyleGAN inversion method II2S \cite{ii2s}. However, we found that when the domain of the $I_{trg}$ (e.g. abstract sketch) is far from the source domain (e.g. FFHQ), the existing inversion model failed to reconstruct the attribute of $I_{trg}$.

  \begin{figure}[!hbt]
\centering
\includegraphics[width=\linewidth]{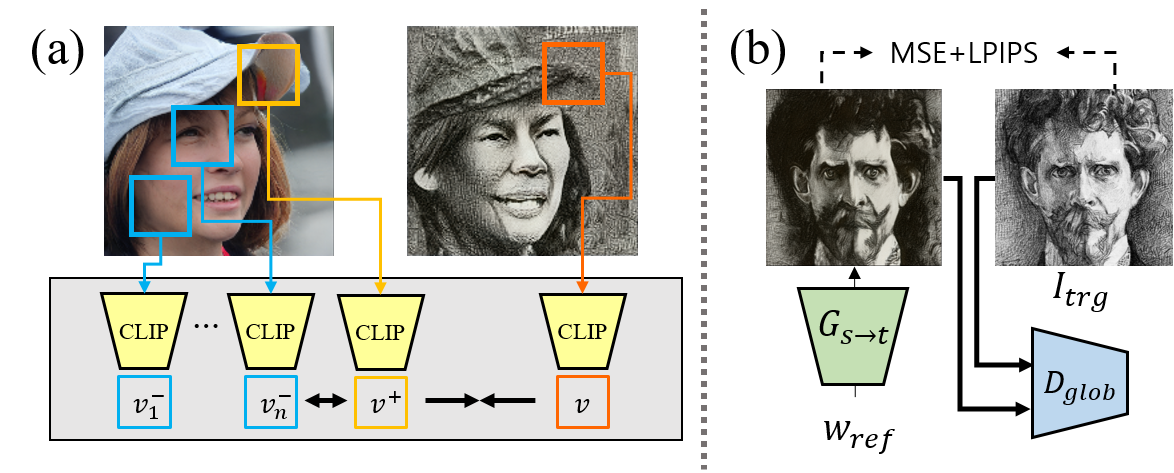}
\caption{(a) Patch-wise semantic consistency in the CLIP space. We crop the patches from generated images and embed the cropped patches using a pre-trained CLIP model. Then we apply contrastive learning on those embedded vectors. (b) Reference target alignment. In the case of using reference latent $w_{ref}$, the generated image should match  $I_{trg}$.  }
\label{fig:method2}
\end{figure}

We conjecture that this performance degradation is caused by the existing inversion model focusing only on pixel-wise similarity. To correct this, we propose to guide the generated $I_{ref}$ to follow the semantic attributes of $I_{trg}$ using the pre-trained CLIP model.
More specifically, as shown in Figure \ref{fig:opt}(left), we include additional loss that reduces the cosine distance between the CLIP space embeddings of the two images while using the pixel-wise loss between the $I_{ref}$ and $I_{trg}$. In addition, we employ the augmentation for $I_{ref}$ to avoid artifacts.  The resulting optimization problem
is formulated as:
\begin{align}
\nonumber
    \arg \min_{w\in W} D_{CLIP}(I_{ref},I_{trg})+ ||I_{ref}-I_{trg}||_2 + \\
    \nonumber
    lpips(I_{ref},I_{trg}) + \lambda_{reg}||w-\Bar{w}||_2,
\end{align}
where $I_{ref}=G_s(w)$, $D_{CLIP}(\cdot,\cdot)$ is cosine distance between CLIP embeddings, $lpips$ is perceptual loss, and
$\Bar{w}$ is {the averaged style code of the source generator $G_s$.}
Here,  we additionally use $l_2$ regularization between $w$ and $\Bar{w}$
  to avoid reconstructing unrealistic images.
Furthermore, to improve the efficiency of the search,
rather than  using random starting point for $w$, we set $\Bar{w}$ as a starting point of $w$.
After optimization step, we use the final latent codes $w$ as the reference latent $w_{ref}$. 


The results of the proposed CLIP-guided optimization are shown in Fig. \ref{fig:opt}(right). When compared to the existing inversion methods II2S \cite{ii2s} and e4e \cite{e4e}, we can observe that the images from our method reflect the attributes (e.g. gender, component shape) of the target image while the baselines fail. 
Since our purpose is to obtain $I_{ref}$ representing the source domain while accurately having the attributes of the target image, the proposed method is more suitable 
in generating a reference.

\subsection{Step 2: Generative Model Fine-tuning}

\subsubsection{Cross-domain semantic consistency} In the next step, we create a target generator $G_t$ by fine-tuning the weights of the pre-trained generator. 
Recall that  Ojha et al \cite{Fewshot} used  a patch discriminator $D_{patch}$ so that the images generated by $G_t$ have the texture of the target image $I_{trg}$. However,  the use of the  discriminator alone easily leads to overfitting, so  it is necessary to use additional regularization so that $G_t$ can inherit the generation diversity of $G_s$. 
To address this, the previous works attempted to  match the distributions  of the features between the source and target generators. 
Unfortunately, under the single-shot condition,  such regularization based on the distribution of generator features does not work and still causes the overfitting.

Similar to Step 1,  we found that CLIP space regularization solves the issues.
Specifically, we found that maintaining the similarity distribution in the CLIP space as in Fig. \ref{fig:method} is more effective in preventing overfitting because it considers the semantic information of the generated images.  Specifically, when arbitrary latent variables $[w_i]^N_0$ are sampled, we first calculate sample-wise similarity scores for $G_s$ and $G_t$ defined as:
\begin{align}
\nonumber
    c^s_n = D_{CLIP}(G_s(w_i),G_s(w_j)), \\
    \nonumber
    c^t_n =  D_{CLIP}(G_t(w_i)),G_t(w_j)),
\end{align}
where $n$ is the reordered index of $(i,j), i\neq j$ and $D_{CLIP}$ denotes
the cosine similarity in the pre-trained CLIP embedding space.
  With the calculated similarity scores,  the similarity loss between  $G_s$ and $G_t$ can be computed. Different from the existing work \cite{Fewshot} which use softmax with KL divergence for distribution matching, we simply used $l_2$ distance between the similarity scores,
   since we observed that the use of KL divergence in the  CLIP space reduces the training stability. Accordingly, our loss function for semantic consistency between
   $G_s$ and $G_t$ is formulated as:
 \begin{align}
     \nonumber
     L_{con}=\mathop{\mathbb{E}}_{w\sim p_w(w)} \sum_{n}||c_n^s-c_n^t||_2 .
 \end{align}

\subsubsection{Patch-wise semantic consistency} 
Although we can avoid the overfitting problem using the novel loss $L_{con}$,  the loss is a regularization for the sample-wise semantics of the generated images so that
 their local features are ignored. Therefore, for better consistency of fine details between source and target generators, we propose a new patch-wise consistency loss. 

To preserve the local attribute between two domains, we start from the idea of patch-wise contrastive loss (PatchNCE) in  CUT \cite{cut}. 
Recall that CUT employs the contrastive learning to the embedded features of the generator. Instead of using generator features directly, additional MLP header network is used in CUT to embed the features into another space. However when we applied this directly to our framework, we observed that the training fails due to the imbalance between the header and the pre-trained generator.
For this problem, we also found that CLIP is essential so that  we propose to use the pre-trained CLIP model as a patch-wise embedding network.

Specifically, after cropping patches at random locations from the images generated by $G_s$ and $G_t$, we embed the image patches with the CLIP encoder as shown in Fig.~\ref{fig:method2}(a).  Then, we reduce the distance between the positive patches cropped at the same location, and push away the negative patches cropped from other locations. 
More specifically, if we set arbitrary location $s_0$, the cropped patch of the outputs from the generators $G_s$ and $G_t$  are denoted as $[G_s(w)]_{s_0}$ and $[G_t(w)]_{s_0}$, respectively. Then if we set other $N$ locations, we can obtain $N$ patches from source domain as $[G_s(w)]_{s_i}$ where $i\in\{1,...,N\}$. With the cropped patches, we can calculate the patch-wise loss such as:
\begin{align}
\nonumber
    L_{patch} = -\text{log}\left[\frac{\text{exp}(v\cdot v^+)}{\text{exp}(v\cdot v^+) + \sum^N_{i=1}\text{exp}(v\cdot v_i^-)}\right],
\end{align}
where $v=E([G_t(w)]_{s_0})$, $v^+=E([G_s(w)]_{s_0})$ are positive vectors, and $v^-=E([G_s(w)]_{s_i})$ are negative vectors. Here, 
$E$ refers to the pre-trained CLIP encoder for the image patch, and the dot mark $\cdot$ represents cosine similarity.

\subsubsection{Reference Target Alignment} 
{The losses suggested in the previous part serves to prevent overfitting when we sample arbitrary latents. This should be alternated with the additional loss
for  generating images through reference latent $w_{ref}$ found in Step 1.  Since $I_{ref}$ obtained in Step 1 has the attribute most similar to $I_{trg}$ and is an image representing the source domain, the domain-adapted output $G_t(w_{ref})$ should be matched to $I_{trg}$. Therefore, as shown in Fig.~\ref{fig:method2}(b), we match the images in both of pixel and perceptual perspective as proposed in \cite{mind}. Also, we further guide the $G_t(w_{ref})$ with global discriminator $D_{glob}$ to make the image much closer to $I_{trg}$.}


\subsubsection{Overall Training}  
Overall, our network is trained by alternately minimizing the two losses. 
First, when we use reference latent $w_{ref}$, our loss is defined as:
\begin{align*}
   \nonumber
    L_{ref} = ||G_t(w_{ref})-I_{trg}||_2+ lpips(G_t(w_{ref}),I_{trg}) \\ 
    \nonumber
    + L^g_{adv}(G_t,D_{glob}),
\end{align*}
where StyleGAN2 adversarial loss is defined as $L^g_{adv}(G,D_{glob}) =  D_{glob}(G(w_{ref})) -D_{glob}(I_{trg})$. For a global discriminator, we fine-tuned the pre-trained StyleGAN2 discriminator.
Second, when arbitrary latents $w$ are sampled, our loss is:
\begin{align}
    \nonumber
    L_{rand} = \lambda_{con}L_{con}+  \lambda_{patch}L_{patch} + L^p_{adv}(G_t,D_{patch}).
\end{align} 
In this case, $L^p_{adv}$ is defined as $L^p_{adv}(G,D) = D_{patch}(G(w))-D_{patch}(I_{trg})$.
 For $D_{patch}$, we used the proposed network in Ojha et al \cite{Fewshot}, {in which $D_{patch}$ is a subset of $D_{glob}$. More specifically, we extract the intermediate feature from $D_{glob}$, then obtain the logit of $D_{patch}$ with mapping the feature through several conv layers. }
Since simultaneously training using two losses is observed to reduce training stability and
  consume excessive memory,
{we therefore alternately train the model 3 iteration only using  $L_{rand}$ (Fig.~\ref{fig:method},~Fig.~\ref{fig:method2}(a))}, then 1 iterations only using $L_{ref}$ (Fig.~\ref{fig:method2}(b)).

\subsection{Text-guided Image Manipulation}

For text-guided image manipulation, we  replaced the adversarial loss of our model with the directional CLIP loss. More specifically, in the first step we find the reference image $I_{ref}$ that most closely matches the target text constraint $t_{trg}$.
Specifically, our goal is  to find latent code $w_{ref}$ for the source domain generator $G_s$ as follows:
\begin{align}
\nonumber
 w_{ref}:=   \arg \min_{w\in W} D_{CLIP}(I_{ref},t_{trg})+ \lambda_{reg}||w-\Bar{w}||_2,
\end{align}
where $I_{ref}=G_s(w)$, 
$D_{CLIP}(\cdot,\cdot)$ is the cosine distance between CLIP embeddings, and $\bar{w}$ is averaged style code of the source generator. After the optimization step, we use the final latent code $w$ as the reference latent $w_{ref}$.

Next, we fine-tune the generator model using the directional CLIP loss proposed in StyleGAN-NADA~\cite{nada}: 
\begin{align}
\nonumber
\Delta T = E_T(t_{trg})-E_T(t_{src}),\\
\nonumber
\Delta I = E_I(G_t(w))-E_I(G_s(w)), \\
\nonumber
L_{dir} = 1-\frac{\Delta I\cdot\Delta T}{|\Delta I||\Delta T|},
\end{align}
where $E_T$ is text encoder of CLIP, and $E_I$ is image encoder of CLIP, $w$ is randomly sampled latent, and $t_{source}$ is the source text condition 
representing the text description of the domain of $G_s$. In our case, we used the text ``Photo" as $t_{src}$. 
In order to prevent overfitting problems, again we use our proposed consistency losses $L_{con}$ and $L_{patch}$.



Our full loss function for $G_t$ fine-tuning is then defined as:
\begin{align}
    \nonumber
    L_{text} = L_{con} + L_{patch} + L_{dir} + L_{dir}^{ref},
\end{align}
where $L_{dir}^{ref}$ is the directional clip loss with using reference latent $w_{ref}$ instead of sampled $w$. 
Due to the absence of target images, we do not use discriminator network. We trained the model for 1,000 iterations.

{Similar to Mind the GAP~\cite{mind}, instead of showing direct output $G_t(w)$,
 we also use  $G_t(w_{mix})$, where the mixed style code $w_{mix}$ was constructed by replacing the last 11 vectors of sampled latent $w$ with $w_{ref}$. }

\begin{figure*}[t!]
\centering
\includegraphics[width=\linewidth]{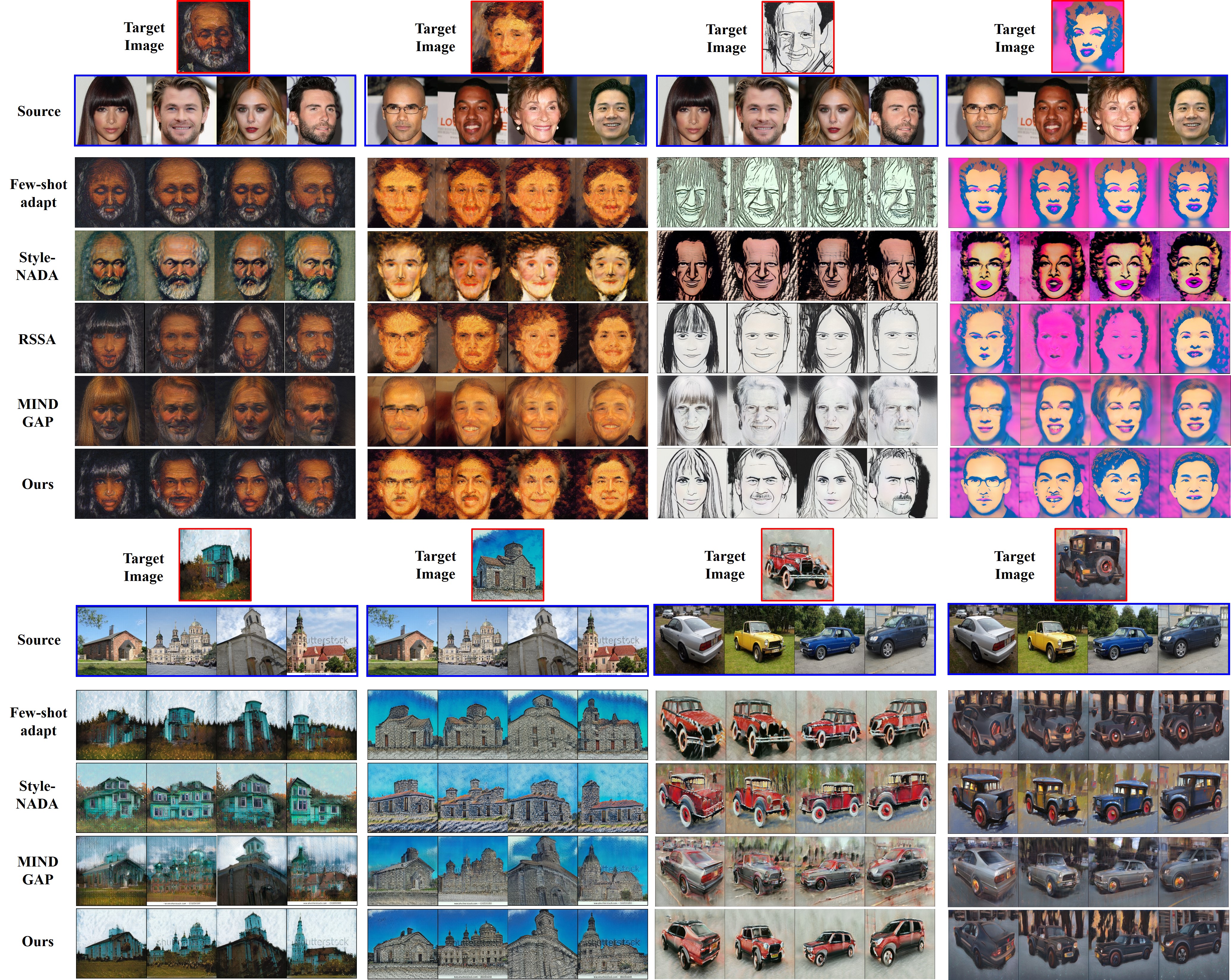}
\caption{Comparison results with baseline models. The images in the uppermost row are the target  images, and the images in the second roe are source content images. The results in the following rows are generated by the models adapted with target images. Our results contain the texture of the target images while preserving the content attributes of the source images.}
\label{fig:result_face}
\end{figure*}

\section{Experiments}
\subsection{Experimental details}
Our training starts with the pre-trained StyleGAN2~\cite{stylegan2}. To verify the versatility of our model, we conducted three different experiments using  pre-trained generators trained on three datasets (FFHQ, LSUN church, LSUN cars). The models trained on FFHQ, LSUN church, and LSUN cars generate 1024x1024, 256x256, and 512x512 resolution images, respectively. We used the Adam optimizer \cite{adam} for all cases, and set the learning rate to 0.02. We used batch size 2 for FFHQ and LSUN cars, and batch size 4 for LSUN church. The training time took about 30 minutes with 2,000 iterations using a single RTX 2080 GPU. This is the time including 3 minutes of CLIP-guided optimization in step 1. {For $L_{patch}$ calculation, we used patch size of 128x128, 64x64, and 32x32 in adapting the pre-trained models on FFHQ , LSUN church, and LSUN cars, respectively. }
For hyperparameters, we set $\lambda_{reg}$ in step 1 as 0.01, and set $\lambda_{con}$, $\lambda_{patch}$ as 10, 1, respectively. Similar to previous work \cite{mind}, we freezed `toRGB' layer and mapping network in the refining stage.

For the patch discriminator $D_{patch}$, we used   a subset of $D_{glob}$ as proposed by Ojha et al.\cite{Fewshot}. More specifically, using a pre-trained discriminator of StyleGAN2, we first extract the intermediate features using the first $l$ residual conv layers of $D_{glob}$, and then map the s to the final logit through an additional conv layer $F^l$(CONV3$\times$3-LeakyReLU). In order to deal with various patch window sizes,  multiple $l$ and $F^l$ were used.
{Specifically,  when we adapt the model pre-trained with FFHQ, we used $l=[5,6]$; for the pretrained model with LSUN church \cite{LSUNchurch}, 
we used $l=[3,4]$; for the pretrained models with LSUN cars~\cite{LSUNcar} and AFHQ dog~\cite{starganv2}, $l=[4,5]$ was used.}

For the CLIP-guided latent optimization, in contrast to the baseline methods such as II2S~\cite{ii2s} and e4e~\cite{e4e}, which use extended StyleGAN latent space $W^+\in \mathbb{R}^{18\times512}$, the optimization was performed in the single latent space in $W\in \mathbb{R}^{512}$ since it produced better perceptual quality and preserved $I_{trg}$ attribute.

{For inference, in the case of adapting FFHQ-pretrained model, we inverted the CelebA-HQ \cite{progan} images into latent codes $w_{rec}$ using e4e \cite{e4e} and  generated images $G_t(w_{rec})$. 
In the case of LSUN church and LSUN cars models, we do not have proper validation image sets. Therefore we showed the results with same sampled latent $w'$, in which the source image is $G_s(w')$ and adapted outputs are $G_t(w')$
}

{With the given latent code $w$, we could show the adapted output with $G_t(w)$. However, we observed that the generated images often have minor artifacts in background. To alleviate them, we used style mixing trick in which we replace the fine layer style code to average style vector $\Bar{w}$ of $G_t$. In case of FFHQ model adaptation, instead of using $w$ which is composed of 18 vectors, we use mixed style code $w_{mix}$ in which the last 7 vectors  are substituted to $\Bar{w}$.
 In the LSUN car model fitting experiment, the last 7 vectors in the sampled code $w$, which consisted of 16 vectors, were replaced; and in adapting the LSUN church model, the last 6 vectors in the sampled code $w$, which consisted of 14 vectors, were replaced. For adapting the model with AFHQ dog, the last 9 vectors in the sampled code $w$, which consisted of 16 vectors, were replaced.
}



For training iterations, our default setting of training iteration is 2,000. However, if the content shape of target image is not much far from the source domain (e.g. realistic portrait), we trained the model with only 1,000 iterations. We selected perceptually better image between the outputs from models trained on 1,000 and 2,000 iterations. In case of adapting LSUN church models, we used 1,500 iterations since the model converged faster than the  others.

For target images,  we selected random image from the portrait face dataset~\cite{portrait} in FFHQ model adaptation. For others, we randomly collected publicly available images from the internet.


\subsection{Qualitative Results}
Fig.~\ref{fig:result_face}(up) shows the image result generated by the target generator adapted from the source generator pre-trained with FFHQ. For comparison, we also show the images generated from the state-of-the-art baseline models: Few-shot GAN adaptation \cite{Fewshot}, StyleGAN-NADA \cite{nada}, Relaxed Spatial Structural Alignment (RSSA)~\cite{rssa}, and Mind the Gap \cite{mind}. In all cases, the results show that the baseline models of few-shot GAN adaptation and StyleGAN-NADA suffer from overfitting problems, and in some cases, the model training for adaptation failed.
For RSSA, the model showed diverse generated outputs which indicates that the model partially solved the overfitting problem. However, the generated outputs still show same facial components (e.g. eyes), which indicate that the model could not fully address the overfitting problem.  

In the case of Mind the Gap, unlike previous baselines, the model generates diverse images while preserving the content of the source domain images. However, the generated images did not properly reflect the domain texture of the target image  in most cases, and only relatively easy characteristics such as global color or background were mainly changed. 
On the other hand, our model generates images that accurately reflect the complex texture (e.g. brushstroke, sketch line) of the target image while preserving the contents of the source domain images. 

In order to verify the versatility of the proposed model, we conducted additional experiments with domain adapted models using source generator pre-trained with LSUN cars and LSUN church datasets. We also show the results from baselines models  for comparison in Figure~\ref{fig:result_face}(down).  Again, we observed that our model perceptually improved resulted compared to other baselines. Since RSSA model mostly concentrates on human face dataset, we did not include the qualitative outputs. 

\begin{figure}[t!]
\centering
\includegraphics[width=\linewidth]{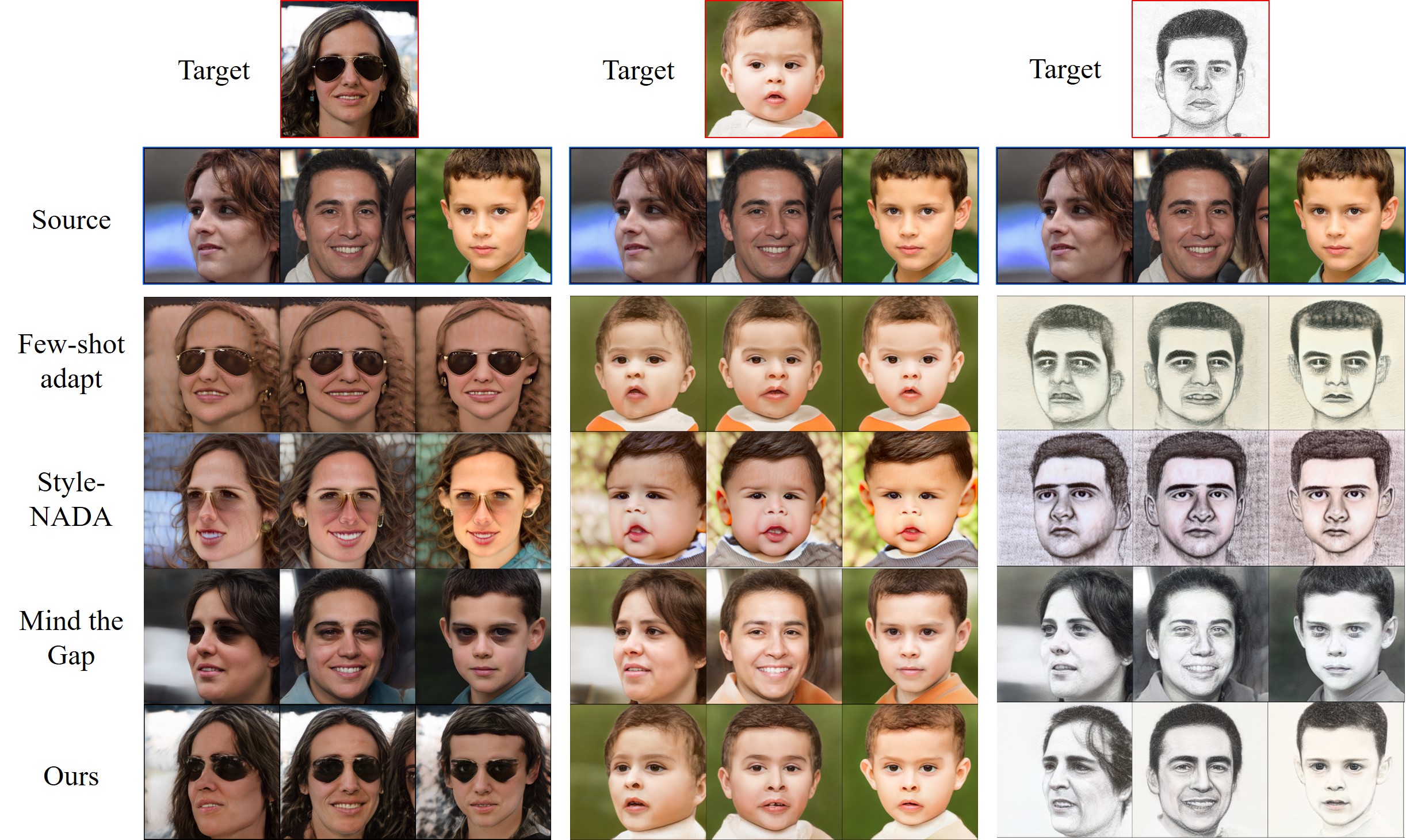}
\caption{{Additional qualitative results on sketches, FFHQ-sunglasses, FFHQ-babies datasets. The result from our model reflects accurate semantic information of the target image while baseline models fail.}}
\label{fig:sketch}
\end{figure}

In Fig.~\ref{fig:sketch}, we show qualitative results from models trained with sketches, sunglasses, and babies datasets. Our model also outperforms baseline methods,  showing successful translation outputs while preserving the content structure of the sources.

\begin{figure}[t!]
\centering
\includegraphics[width=\linewidth]{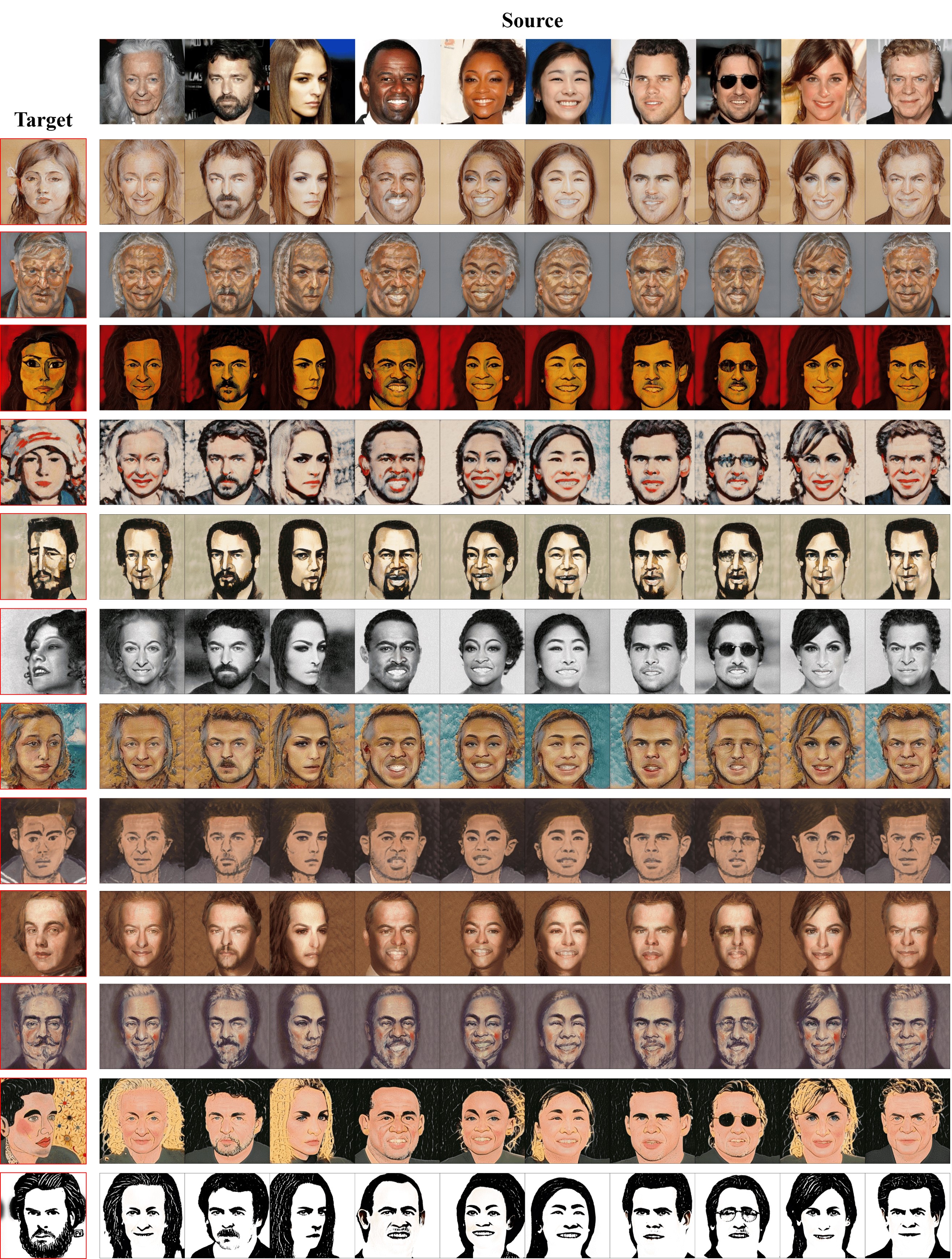}
\caption{Additional results with adapting the models pre-trained on FFHQ dataset. The images in the uppermost row are the source content images. The  results in the following rows are generated by the models adapted with the  target images.}
\label{fig:result_add1}
\end{figure}

\begin{figure}[t!]
\centering
\includegraphics[width=1.0\linewidth]{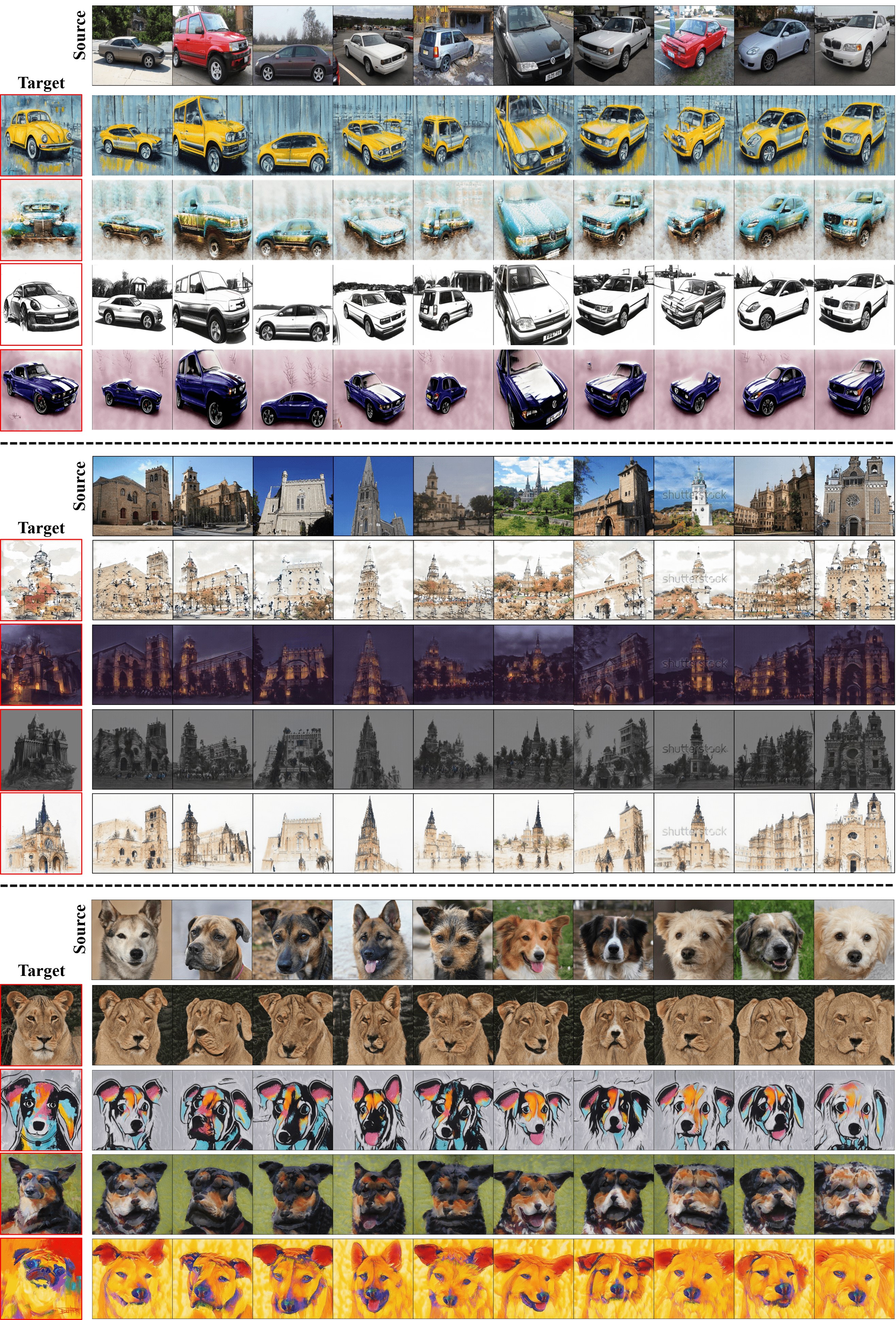}
\caption{Additional results with adapting the models pre-trained on LSUN Cars (Up), LSUN church (Middle), and AFHQ Dog (Bottom) datasets. The images in the uppermost row are the source content images. The  results in the following rows are generated by the models adapted with the target images.}
\label{fig:result_add2}
\end{figure}

For further evaluation, we show additional results on adapting FFHQ pre-trained models in Fig.~\ref{fig:result_add1}. In Fig.~\ref{fig:result_add2}, we also show the results on adapting the models pre-trained on LSUN Car, LSUN church, and AFHQ dog datasets. The generated images have semantic style of the target images while preserving the content information of source images.

\subsection{Quantitative Results}
\begin{table}[!t]
\centering
		\begin{tabular}[b]{c|ccc|cc}
		
        {\textbf{Metrics}}&\multicolumn{3}{c|}{\textbf{FID$\downarrow$}}&\textbf{LPIPS$\uparrow$}&\textbf{ID$\uparrow$}\\
        \hline
       {\textbf{Dataset}}& \textbf{Sketches} & \textbf{Sunglasses} & \textbf{Babies} & \multicolumn{2}{c}{\textbf{Portrait}} \\ \hline
       TGAN &116.8 & 96.92 & 152.7&0.336& 0.100\\
       Mine & 117.9 & 85.26 &168.2 &0.413& 0.150\\
       FD & 128.5& 108.1 &176.3 &0.392&0.146\\
       
      FSA & 145.8& 90.34 &119.6 &0.437&0.163\\
      NADA & 154.8& 137.8 & \textcolor{red}{102.7} &0.455& 0.175\\
      MTG & \textcolor{blue}{107.2}&\textcolor{blue}{77.34} & 123.6&{0.554}&{0.343}\\
      RSSA & 112.7& 108.4& 123.8& \textcolor{blue}{0.558} &  \textcolor{blue}{0.396}\\
      Ours & \textcolor{red}{83.87}&\textcolor{red}{64.61} & \textcolor{blue}{105.2} &\textcolor{red}{0.568}& \textcolor{red}{0.405}\\
    \hline
    \end{tabular}
	\captionof{table}{Quantitative results on various metrics of FID, LPIPS, and identity score \textcolor{blue}{Blue-second best}, \textcolor{red}{Red-best}}
	\label{tab:quant}
\end{table}

\begin{table}[t]
\centering
\begin{tabular}{@{\extracolsep{5pt}}c|cccc@{}}


{\textbf{Methods}} &FSA & NADA & MTG & Ours\\


\hline
Score $\uparrow$ & 1.96 & 1.83 & 3.37 & \textbf{3.92} \\
\hline

\end{tabular}
\caption{User study results on various single-shot domain adaptation models. Our model outperforms baseline methods in user study.  }

\label{table:user}
\end{table}













For evaluation of our proposed model, we conducted quantitative experiments in Table~\ref{tab:quant}. First, we measured the generation quality with FID~\cite{fid} scores. For fair comparison, we performed adaptation for sketches~\cite{sketch}, FFHQ-sunglasses, and -babies data~\cite{stylegan}, which are benchmarks for GAN adaptation. To experiment with the one-shot setting, we randomly selected 5 images from each dataset and used them as targets. With each adapted model, we randomly generated 1,000 samples and calculated FID between the real and generated data. We report average score of 2 training runs for each target image (total 10 runs for each score). In the experiments for FFHQ-babies and sunglasses, we used smaller parameters ($\lambda_{con}=0.1,\lambda_{patch}=0.01$) as it showed better performance.

In order to further evaluate the generation diversity, we also calculated intra-sample LPIPS distance~\cite{lpips}. For each adapted model, we randomly generate two batches and calculated LPIPS distance between the batches. We used averaged score of 1,000 random batch pairs, and each batch contains 4 samples. We experimented on the models shown in our main paper, which are adapted on portrait painting dataset. We selected 10 different models for each method and report the averaged scores.

To evaluate the content preservation performance, we report the face identity score~\cite{arcface} which are calculated between the images generated from both models $G_s(z)$ and $G_t(z)$, in which $z$ are the same latent codes. 

As baselines, we adopted several few-shot adaptation models such as TransferGAN (TGAN)~\cite{TGAN}, FreezeD (FD)~\cite{freezeD}, MineGAN (Mine)~\cite{mine}, Few-shot adaptation (FSA)~\cite{Fewshot}, StyleGAN-NADA (NADA)~\cite{nada}, Relaxed Spatial Structural Alignment (RSSA)~\cite{rssa}, and Mind the Gap (MTG)~\cite{mind}. The quantitative results show that our model obtained the best FID score in sketches and sunglasses datasets, and the second best in the babies dataset. Our model also scored the best in LPIPS generation diversity and identity scores.
The results show that our model shows the best performance among all baselines, with respect to generation quality, diversity, and content preservation. 



\subsection{User Study}
For further evaluation of our proposed model, we additionally conducted a user study in Table~\ref{table:user}. To quantitatively measure the detailed preference from users, we used a custom-made mean opinion scoring system.  
As baselines, we used the models of Few-shot adaptation, StyleGAN-NADA, and Mind the Gap. 

  For user study,  we provided the generated images to users, and asked them to score the images according to three evaluation criteria: \textcolor{red}{1)} Do the generated images properly reflect the domain style of the target image? \textcolor{red}{2)} Does the content of the generated images properly match with the source domain images? \textcolor{red}{3)} Do the generated images have sufficient diversity? Users can choose the scores among 5 options: 1-very bad, 2-bad, 3-neutral, 4-good, 5-very good. We randomly recruited 30 users using Google Form, who come from the age group between 20s and 40s. We provided users with 100 generated images per model (total 400 images). 

 In Table~\ref{table:user},  we can see that our model outperforms all baseline models in  terms of perceptual preference scores.

\subsection{Text-guided Image Manipulation Results}

\begin{figure}[t!]
\centering
\includegraphics[width=\linewidth]{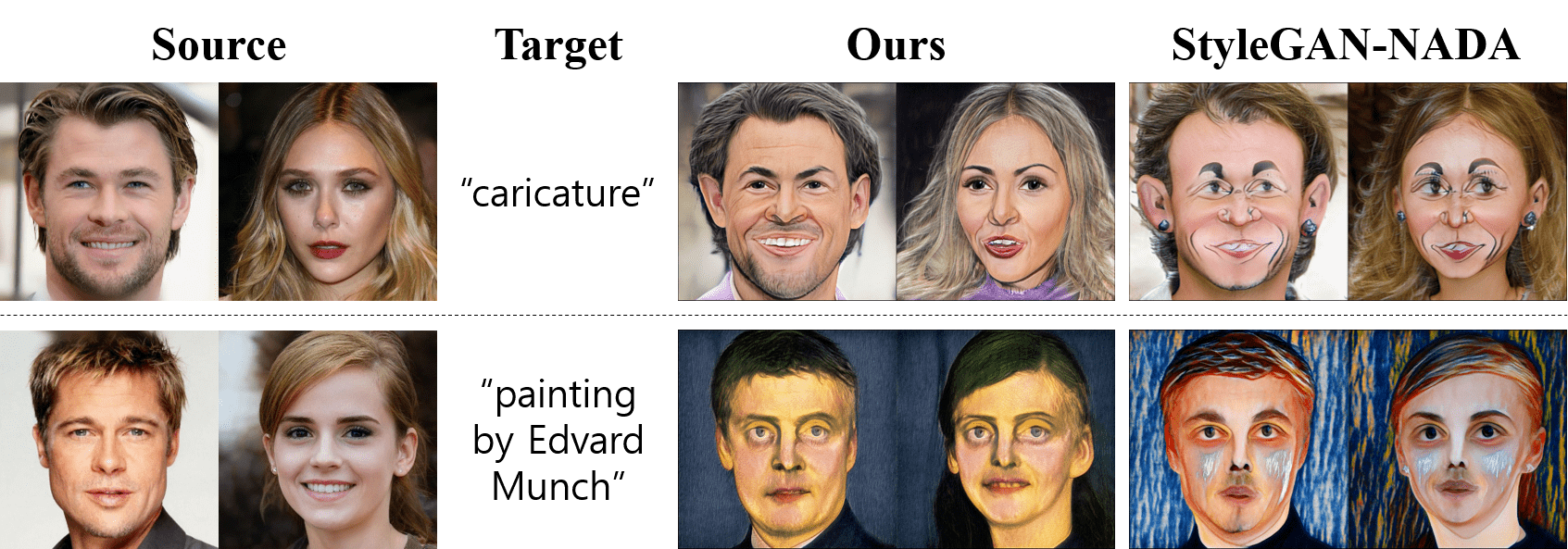}
\caption{Text-guided adaptation results. The results from our model follow the overall style of text condition with better content preservation compared to StyleGAN-NADA. }
\label{fig:text}
\end{figure}

\begin{figure}[t!]
\centering
\includegraphics[width=1.0\linewidth]{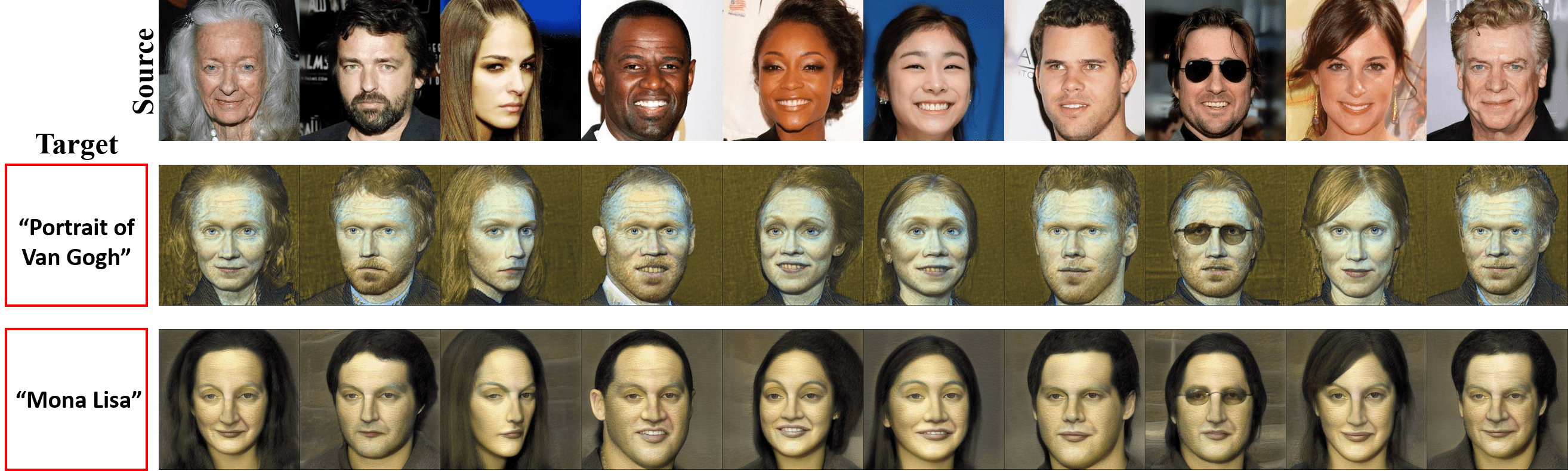}
\caption{Additional results from our  text-guided model adaptation. The images in the uppermost row are the source content images. The  results in the following rows are generated by the models adapted given the target texts at the leftmost column.}
\label{fig:result_supp_text}
\end{figure}

\begin{table}[t]
\centering
\begin{tabular}{@{\extracolsep{5pt}}c|cccc@{}}

{\textbf{Methods}} &Style-NADA & Ours\\


\hline
Score $\uparrow$ & 2.46 &\textbf{3.64} \\
\hline

\end{tabular}
\caption{User study results on text-guided manipulation models. Our model outperforms baseline StyleGAN-NADA in user study.  }
\label{table:user_text}
\end{table}

In Fig.~\ref{fig:text}, we show comparison results of model adaptation with single text condition. Although baseline StyleGAN-NADA can modulate the overall texture of outputs, the generated faces have almost same identity regardless of the input images,  suggesting that  the model suffers from overfitting. In contrast, our model can synthesize reasonable outputs that reflect both  text condition and source content. The results showed that our CLIP space regularization  can avoid overfitting,  further confirming that our proposed model have a strong advantage in versatility. 

To further show the superiority of our text-guided image manipulation model, we conducted another user study. We followed the same protocol used in our one-shot adaptation experiments. Again, we provided the generated images to users, and asked them to score the images according to three evaluation criteria: \textcolor{red}{1)} Do the generated images properly reflect the domain style of the target text? \textcolor{red}{2)} Does the content of the generated images properly match with the source domain images? \textcolor{red}{3)} Do the generated images have sufficient diversity? 

Users can choose the scores among 5 options: 1-very bad, 2-bad, 3-neutral, 4-good, 5-very good. We randomly recruited 20 users using Google Form, who come from the age group between 20s and 40s. We provided users with 40 generated images per model, which are generated from 4 different text conditions (e.g. ``cacicatures"). We compared our model with StyleGAN-NADA.
Table~\ref{table:user_text} shows that our text-guided model outperforms StyleGAN-NADA in human perception scores.

We further show the adaptation result on text conditions in Fig.~\ref{fig:result_supp_text}.

\subsection{Latent Space Editing}

Recall that our purpose is to train a fine-tuned model $G_t$ to generate images with the target texture while preserving the latent space characteristics of the source domain. To verify this, we tested whether the latent space of $G_t$ sufficiently includes the disentangled characteristic of the source domain through attribute editing. 

In this part, we adopted the latent space editing technique in  StyleCLIP \cite{styleclip}, which is a state-of-the-art attribute editing method. With StyleCLIP, we can edit the face attributes with various text conditions. Among the three editing approaches proposed in StyleCLIP, we used the editing through global direction which can be universally applied. When there is a latent code $w$ corresponding to the source image, we can manipulate it with StyleCLIP to get a edited latent $\hat{w}$, then we put the code as an input to the fine-tuned generator $G_t$ to get the attribute edited output in the target domain. 

\begin{figure}[t!]
\centering
\includegraphics[width=\linewidth]{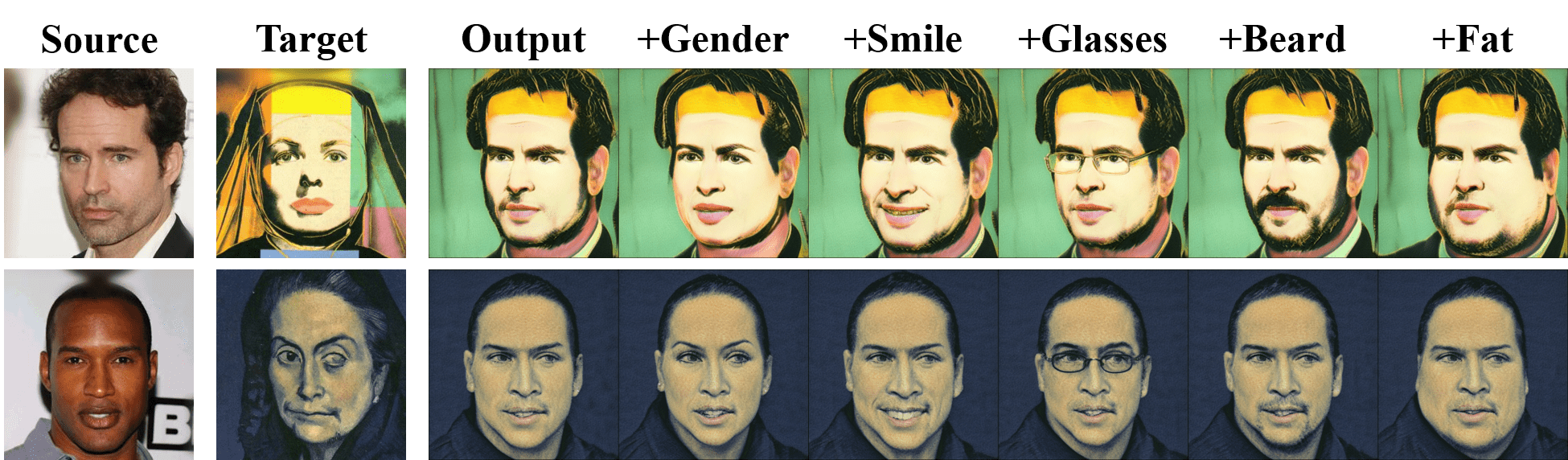}
\caption{Latent space editing results. We manipulated the attribute of the faces using a tool in the StyleCLIP. The results show that adapted model $G_t$ still maintains the latent space characteristic of source domain generator. }
\label{fig:edit}
\end{figure}

We show experimental  results of attribute editing in Fig.~\ref{fig:edit}. When editing through various text conditions, we show that domain-adapted generator $G_t$ can also generate edited outputs without entanglement. This shows that our adapted generator maintains the content characteristics of the source domain.

\subsection{Ablation study}
\begin{figure}[t!]
\centering
\includegraphics[width=\linewidth]{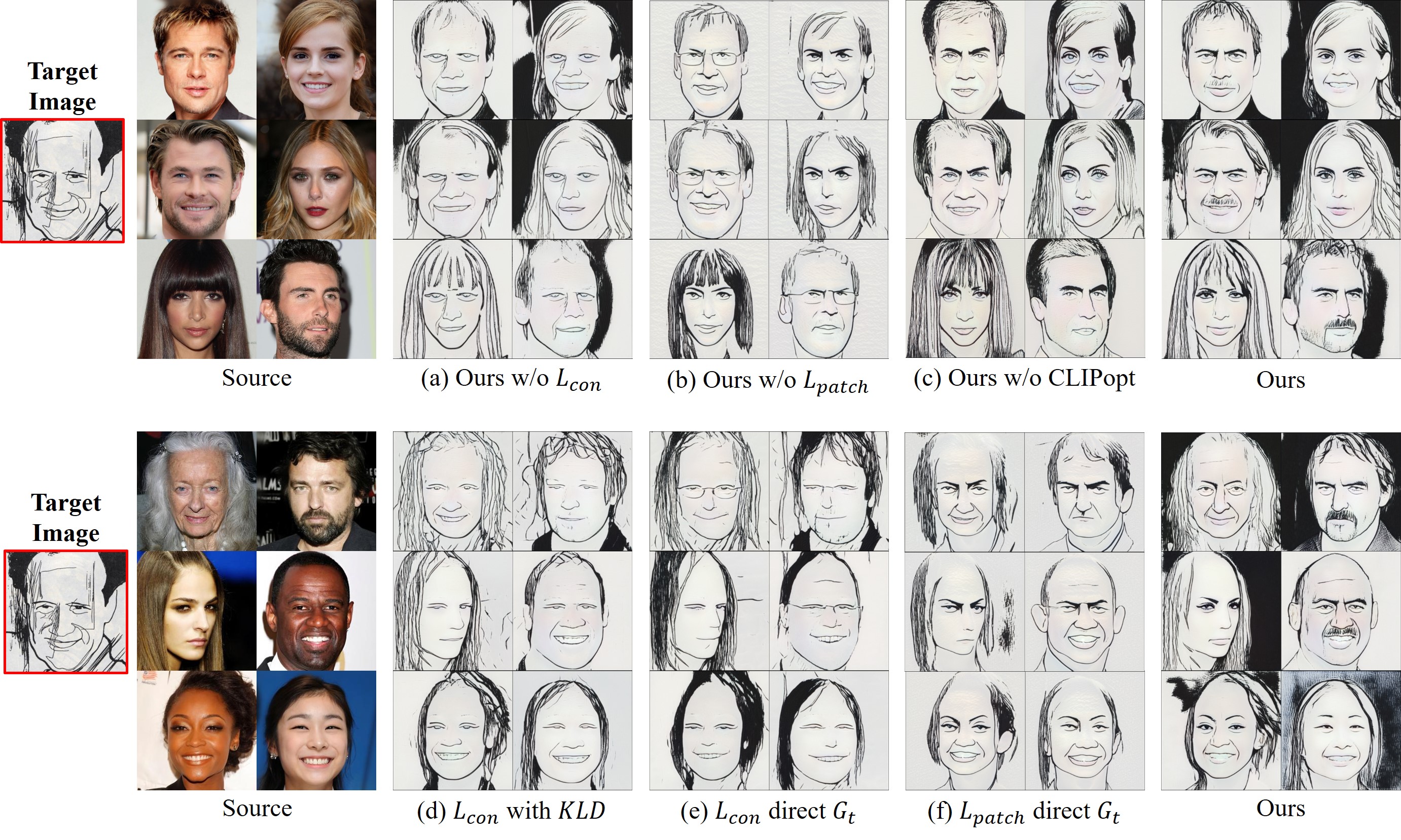}
\caption{Ablations study results.  (a)The absence of $L_{con}$  leads to overfitting, generating the same face identity regardless of the input images. (b) Without $L_{patch}$, detailed component shapes are misaligned with the source image. (c) When  II2S is used for reference research instead of our proposed CLIP guided optimization (CLIP{opt}), several facial attributes are not correctly reflected. (d) Generated images from $G_t$ adapted using $L_{con}$ with Softmax and KL divergence. (e) Generated images from $G_t$ adapted using consistency loss on generator features \cite{Fewshot} instead of our proposed CLIP space consistency $L_{con}$. (f) Results from $G_t$ adapted using contrastive regularization proposed in CUT \cite{cut} instead of our $L_{patch}$. }
\label{fig:ablation}
\end{figure}
		
       

\subsubsection{Qualitative Comparison}
In order to check how the various components proposed in our method operate, we conducted ablation studies by removing each component in Fig.~\ref{fig:ablation}. 

Specifically, \textcolor{blue}{(a)} when our proposed semantic consistency loss $L_{con}$ is removed, we can see that the facial attributes of the generated faces have identical shape due to the overfitting of adapted model. 

 \textcolor{blue}{(b)} When removing the patch-wise consistency loss $L_{patch}$, we can see the generated faces have different face identities from the source domains images. Although using only the suggested $L_{con}$ can prevent overfitting,  we found that when $L_{patch}$ is not used, the generated images are excessively deformed due to lack of consistency between the local regions of source and output images. 

\textcolor{blue}{(c)} To evaluate the superiority of our proposed CLIP-guided optimization, we trained the model using baseline II2S \cite{ii2s} instead of our the proposed CLIP space
optimization. At this time, the model generates decent result images, but we can see that there are misalignments between the attributes (e.g eyes shape, beard) of the generated and source images. We conjecture that the problem occurs because the reference image generated with II2S does not properly cover the attribute of the target image. 

\textcolor{blue}{(d)} When we use $L_{con}$ composed of Softmax \& KL Divergence instead of our proposed $L_2$ regression, we observed that the results suffer from severe overfitting as the loss does not work properly. 

\textcolor{blue}{(e)} Next, when we use the consistency loss on generator features proposed by Ojha et al \cite{Fewshot} instead of our CLIP-space consistency loss $L_{con}$,  the model suffers from severe overfitting due to training failure. 

\textcolor{blue}{(f)}  
We show the results using contrastive regularization on generator features with additional header network $F$ similar to CUT\cite{cut}, instead of using our $L_{patch}$. In this case, we observed that overfitting problem is resolved as $L_{con}$ correctly operates, but the local features of output images are not properly matched to the source images, due to training imbalance between generator and header networks. We can see that using our full settings can preserve the attribute of source images the best.

\subsubsection{Quantitative Comparison}
In Table~\ref{tab:ablation}, we show quantitative ablation study results for our main proposed components.  We measured various quantitative metrics with adapted models on sketches dataset. We followed the same experimental settings of our main quantitative experiments. For additional study, we include two different ablation settings: \textcolor{blue}{(g)}   To show the effectiveness of reference-target matching, we experimented with removing the matching between $I_{ref}$ and $I_{trg}$. 
\textcolor{blue}{(h)}   To further evaluate our proposed mixing trick, we measured the performance with removing the mixing trick.
When the proposed components are removed from our full settings, the models showed degraded performance in all aspects including generation quality (FID), generation diversity (LPIPS), and content preservation (ID). In contrast, our full settings showed the best quantitative performance among all baseline settings.

\begin{table}[!t]
\centering
		\begin{tabular}[b]{c|ccc}
		
        {\textbf{Metrics}}&{\textbf{FID$\downarrow$}}&\textbf{LPIPS$\uparrow$}&\textbf{ID$\uparrow$}\\
        \hline
        w/o Mixing reg \textcolor{blue}{(h)}&90.88 & 0.554 & 0.432 \\ 
        w/o $I_{ref}$ \textcolor{blue}{(g)}&88.83 & 0.537 & 0.394 \\ 
        $L_{patch}$ with CUT \textcolor{blue}{(f)}&115.16 & 0.554 & 0.415 \\ 
        $L_{con}$ with Ojha et al. \textcolor{blue}{(e)} &87.90 & 0.458 & 0.211  \\
        $L_{con}$ with KLD \textcolor{blue}{(d)} &89.49 & 0.482 & 0.295 \\
        
          w/o CLIPopt \textcolor{blue}{(c)}& 92.50 &0.549 & 0.412\\
       w/o $L_{patch}$ \textcolor{blue}{(b)}& 98.34 & 0.536 & 0.436 \\
       w/o $L_{con}$ \textcolor{blue}{(a)}&89.77 & 0.472 & 0.266 \\

      Ours & \textbf{83.87}&\textbf{0.572} & \textbf{0.443}\\
    \hline
    \end{tabular}
	\captionof{table}{Ablation study results on various metrics of FID, LPIPS, and identity score. Our best setting showed the best quantitative score.}
	\label{tab:ablation}
\end{table}

\subsection{Multi-shot Adaptation Results}

For further evaluation of our proposed regularization frameworks, we conducted additional experiments on multiple-shot adaptation task. In Table \ref{tab:multi}, the results show that our proposed model outperformed baseline models in multiple shot adaptation task. Our model showed the best score in all of the metrics including FID, LPIPS diversity, and identity scores. 

\begin{table}[!t]
\centering
		\begin{tabular}[b]{c|ccc}
		
        \textbf{Metrics}&\textbf{FID$\downarrow$}&\textbf{LPIPS$\uparrow$}&\textbf{ID$\uparrow$}\\
        \hline
       
      FSA & 78.15& 0.488 &0.229 \\
      MTG & {101.2}&{0.565} & 0.326\\
      Ours & \textbf{70.13}&\textbf{0.567} & \textbf{0.410} \\
    \hline
    \end{tabular}
	\captionof{table}{Quantitative results on 3-shot adaptation tasks. For evaluation, we use sketches dataset. Our model outperformed few-shot adaptation baselines.}
	\label{tab:multi}
\end{table}
\subsection{Comparison with Style Transfer Methods}
\begin{figure}[t!]
\centering
\includegraphics[width=\linewidth]{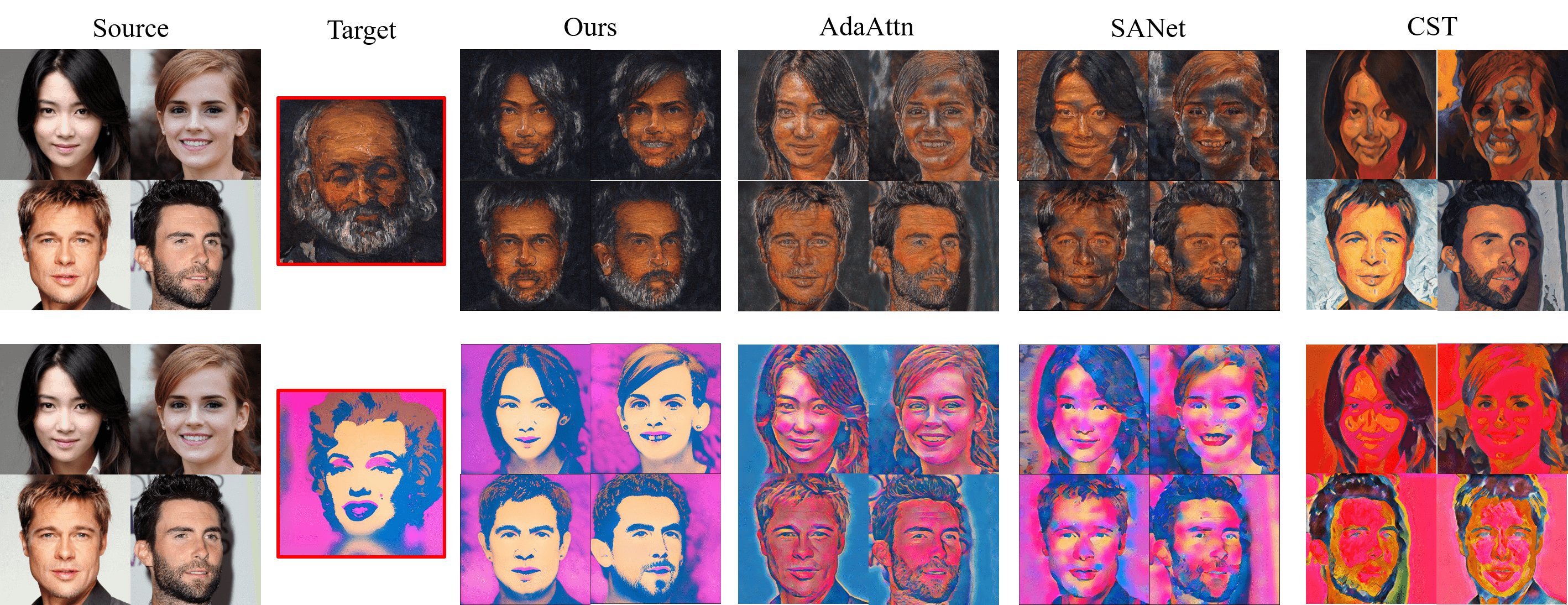}
\caption{{Additional comparison results with state-of-the-art image style transfer models. The result from our model reflects accurate semantic information of the target image while baseline models fail.}}
\label{fig:result_style}
\end{figure}

To further verify the performance of our proposed model, in Fig.~\ref{fig:result_style} we show the comparison results with the existing state-of-the-art image style transfer models, such as
AdaAttn~\cite{liu}, SANet~\cite{park}, and CST\cite{svoboda}.  

In the case of the existing style transfer models, the model failed to reflect the semantic information of target images, as they excessively concentrate on applying the overall color and texture of the target images. We can see that there are several artifacts such as the background color of the target image applied to the face, or the background color applied to entire image areas as shown in Fig.~\ref{fig:result_style}. In the case of our model, we can see that the generated images accurately reflect the characteristics of target domain by considering both the texture and semantic information of the targets.

\section{Discussion and Conclusion}

In this article, we proposed a new framework that can transform a pre-trained StyleGAN to generate target domain images by fine-tuning with a single target image.
The key idea is the CLIP space manipulation through a two step approach. Specifically,
we proposed an optimization method in CLIP space to find the reference image which have the most similar attribute to the target image in the source domain.
The reference image was then used as an anchor point to maintain the content attribute of target generator. In addition,
 to prevent model overfitting, we proposed two different regularization losses: semantic consistency loss and patch-wise consistency both in CLIP space. Our experimental results showed that the proposed method leads to better quantitative and qualitative results than the existing methods. 
 
 {
Our model showed superior performance compared to the baseline models in one-shot domain adaptation experiments, and the results on various adapted models further confirmed that our model has enough versatility. 
As a further extension, we are planning to try domain adaptation on generative models for natural images. In addition, we plan to explore whether our proposed regularization can be applied to other generative framework such as diffusion models.}



\ifCLASSOPTIONcaptionsoff
  \newpage
\fi



%



\bibliographystyle{IEEEtran}
\bibliography{mybib}

\begin{IEEEbiography}[{\includegraphics[width=1in,height=1.25in,clip,keepaspectratio]{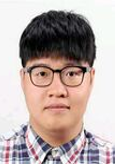}}]{Gihyun Kwon}

received M.S. degree from the Department of Electric Engineering, Korea Advanced Institute of Science and Technology(KAIST), Daejeon, South Korea in 2020, and  
received B.S. degree from the Department of Electronic Engineering, Hanyang University, Seoul, South Korea, in 2018. Currently he is pursuing the Ph.D. degree at the Department of Bio and Brain Engineering, Korea Advanced Institute of Science and Technology. His research interests include machine learning for image processing, especially for generative models. He authored and coauthored multiple papers in top conference proceedings including CVPR, ICCV, MICCAI, and BMVC. 
\end{IEEEbiography}

\begin{IEEEbiography}[{\includegraphics[width=1in,height=1.25in,clip,keepaspectratio]{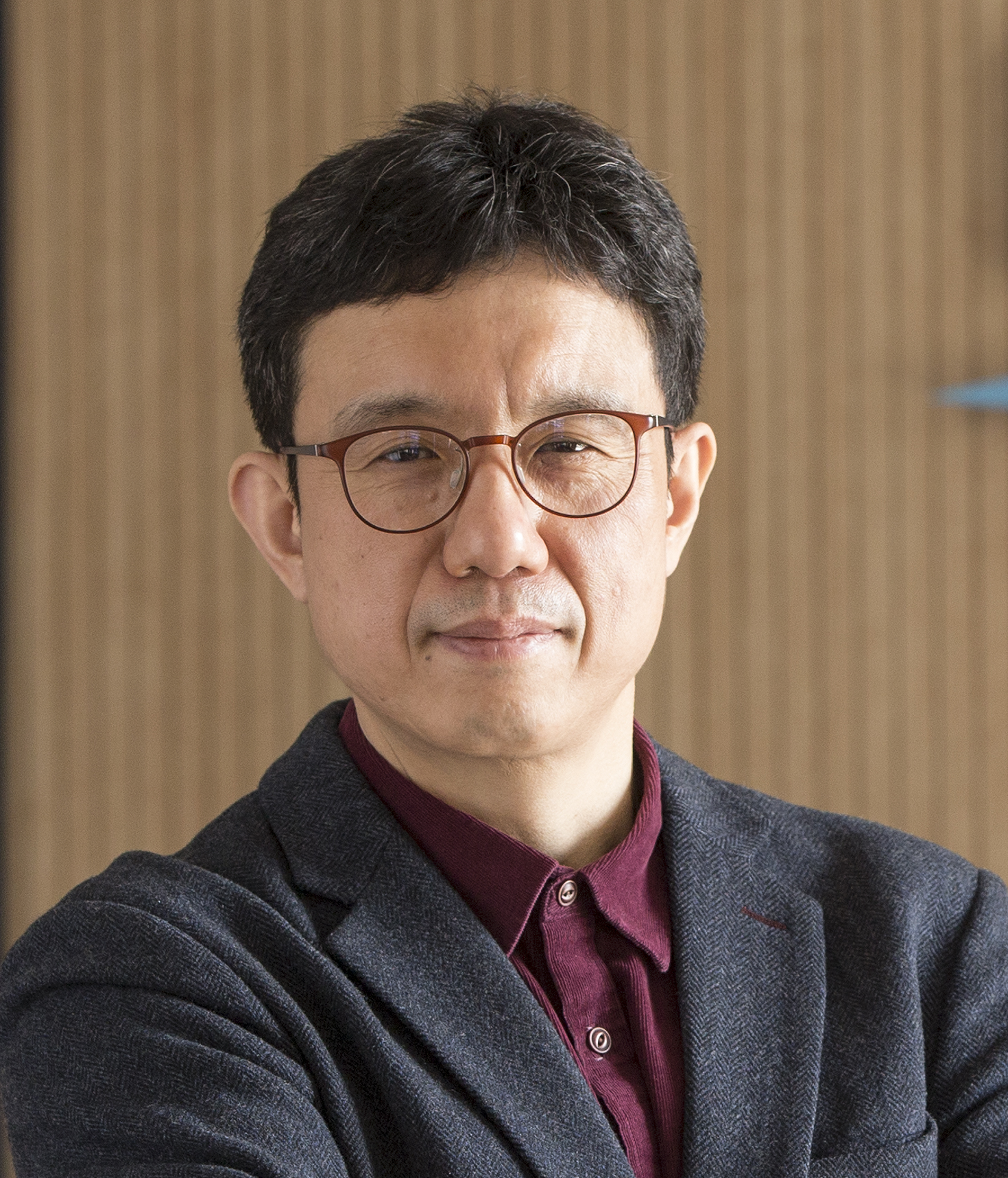}}]{Jong Chul Ye}
(Fellow, IEEE) is a Professor of the Graduate School of Artificial Intelligence (AI) of Korea Advanced Institute of Science and Technology (KAIST), Korea. He received the B.Sc. and M.Sc. degrees from Seoul National University, Korea, and the Ph.D. from Purdue University, West Lafayette. Before joining KAIST, he worked at Philips Research and GE Global Research in New York. He has served as an associate editor of IEEE Trans. on Image Processing, and an editorial board member for Magnetic Resonance in Medicine. He is currently an associate editor for IEEE Trans. on Medical Imaging,  a Senior Editor of IEEE Signal Processing Magazine, and an Executive Editor of Biological Imaging. He is an IEEE Fellow, was the Chair of IEEE SPS Computational Imaging TC,  and IEEE EMBS Distinguished Lecturer. He was a General Co-Chair (with Mathews Jacob) for IEEE Symp. On Biomedical Imaging (ISBI) 2020, and will be a Program Chair for 2024 IEEE International Conference on Acoustics, Speech and Signal Processing (ICASSP 2024). He is also the President of Korean Society of Artificial Intelligence in Medicine (KoSAIM), and the Director of KAIST Center for Digital Health Innovation. His research interest is in machine learning applications and theory for biomedical imaging and computer vision.
%
%
\end{IEEEbiography}

%








\end{document}